\documentclass[10pt,twocolumn,letterpaper]{article}

\usepackage{cvpr}              

\usepackage{microtype}

\usepackage{graphicx}
\usepackage{multirow}
\usepackage{adjustbox}
\usepackage{colortbl}
\usepackage{pifont}
\usepackage{multirow}
\usepackage{makecell}
\usepackage{rotating}
\usepackage{array}
\usepackage[dvipsnames]{xcolor}
\usepackage{subcaption}

\newcommand{\OurMethod}{\emph{OmniMotion-X}}

\newcommand{\TrainingData}{\emph{OmniMoCap-X}}

\definecolor{cvprblue}{rgb}{0.21,0.49,0.74}
\usepackage[pagebackref,breaklinks,colorlinks,allcolors=cvprblue]{hyperref}
\usepackage{bibentry}
\def\paperID{} 
\def\confName{CVPR}
\def\confYear{2026}

\title{OmniMotion-X: Versatile Multimodal Whole-Body Motion Generation}
\author{
    Guowei Xu\textsuperscript{\rm 1}$^{*}$, 
    Yuxuan Bian\textsuperscript{\rm 2}$^{*}$, 
    Ailing Zeng\textsuperscript{\rm 5}$^{\dagger}$, 
    Zhuo Chen\textsuperscript{\rm 1},
    Mingyi Shi\textsuperscript{\rm 3}, 
    Shaoli Huang\textsuperscript{\rm 4}, \\
    Wen Li\textsuperscript{\rm 1}$^{\dagger}$, 
    Lixin Duan\textsuperscript{\rm 1},
    Qiang Xu\textsuperscript{\rm 2}
    \\
    \textsuperscript{\rm 1}UESTC
    \textsuperscript{\rm 2}CUHK 
    \textsuperscript{\rm 3}HKU
    \textsuperscript{\rm 4}Tencent
    \textsuperscript{\rm 5}Independent Researcher 
    \\
    {\tt \{xuguowei368, yuxuanbian23, ailingzengzzz, liwenbnu, lxduan\}@gmail.com}
}

\newcommand{\blfootnote}[1]{%
  \begingroup
  \renewcommand\thefootnote{}\footnote{#1}%
  \addtocounter{footnote}{-1}%
  \endgroup
}

\begin{document}

\twocolumn[{
\renewcommand\twocolumn[1][]{#1}
\maketitle
\vspace{-1.09cm}
\begin{center}
    \captionsetup{type=figure}
    \includegraphics[width=\textwidth]{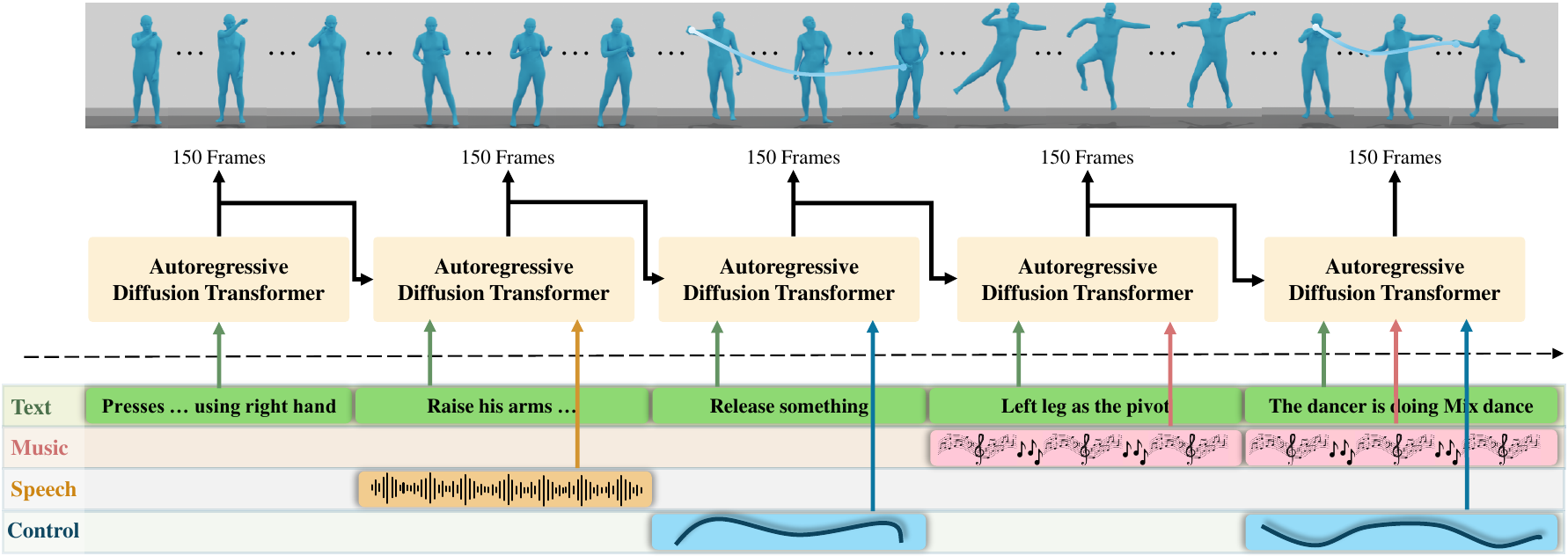}
    \caption{We present \OurMethod, a unified sequence-to-sequence autoregressive motion diffusion transformer designed for flexible and interactive whole-body human motion generation. It supports a variety of tasks, including text-to-motion, music-to-dance, speech-to-gesture, and globally spatial-temporal controllable motion generation, which encompasses motion prediction, in-betweening, completion, and joint/trajectory-guided synthesis. These conditions can be combined in various ways to enable versatile motion generation.}
    \label{fig:introduction}
\end{center}
}]
\blfootnote{$^{*}$Equal contribution.\quad$^{\dagger}$Corresponding authors.}

\begin{abstract}
This paper introduces \textbf{OmniMotion-X}, a versatile multimodal framework for whole-body human motion generation, leveraging an autoregressive diffusion transformer in a unified sequence-to-sequence manner. OmniMotion-X efficiently supports diverse multimodal tasks, including text-to-motion, music-to-dance, speech-to-gesture, and global spatial-temporal control scenarios (e.g., motion prediction, in-betweening, completion, and joint/trajectory-guided synthesis)—as well as flexible combinations of these tasks. Specifically, we propose the use of reference motion as a novel conditioning signal, substantially enhancing the consistency of generated content, style, and temporal dynamics crucial for realistic animations. To handle multimodal conflicts, we introduce a progressive weak-to-strong mixed conditions training strategy. To enable high-quality multimodal training, we construct \textbf{OmniMoCap-X}, the largest unified multimodal motion dataset to date, integrating 28 publicly available MoCap sources across 10 distinct tasks, standardized to the SMPL-X format at 30 fps. To ensure detailed and consistent annotations, we render sequences into videos and use GPT-4o to automatically generate structured and hierarchical captions, capturing both low-level actions and high-level semantics. Extensive experimental evaluations confirm that OmniMotion-X significantly surpasses existing methods, demonstrating state-of-the-art performance across multiple multimodal tasks and enabling the interactive generation of realistic, coherent, and controllable long-duration motions. The code and dataset are publicly available at \url{https://github.com/GuoweiXu368/OmniMocap-X}.
\end{abstract}    
\section{Introduction}\label{sec-1-intro}

\begin{table*}[!t]
  \centering
  \resizebox{0.98\textwidth}{!}{ 
    \begin{tabular}{cccccccccccc}
      \toprule
        \multicolumn{1}{c}{\multirow{2}{*}{Model Type}} & \multicolumn{1}{c}{\multirow{2}{*}{Method}} & \multicolumn{5}{c}{Tasks} & \multicolumn{1}{c}{Reference} & \multicolumn{1}{c}{\multirow{2}{*}{Mixed-condition}} & \multicolumn{1}{c}{\multirow{2}{*}{Whole-Body}} 
        & \multicolumn{2}{c}{Training Data}    \\
        \cmidrule(lr){3-7} \cmidrule(lr){11-12} 
        & & T2M & M2D & S2G & GSTC(S) & GSTC(D) & Motion &  & & Datasets & Hours  \\
      \midrule
      DiT & MDM~\cite{mdm}   & $\checkmark$ & $\times$ & $\times$ & $\times$ & $\times$ & $\times$ & $\times$ & $\times$  & 2   & 28.6 \\
      DiT & MCM~\cite{mcm}   & $\checkmark$ & $\checkmark$ & $\checkmark$ & $\times$ & $\times$ & $\times$ & $\times$ & $\times$  & 3   & 109.8 \\
      DiT & LMM~\cite{zhang2024large}   & $\checkmark$ & $\checkmark$ & $\checkmark$ & $\times$ & $\times$ & $\times$ & $\checkmark$ & $\checkmark$  & 16   & - \\
      DiT & MotionCraft~\cite{bian2024motioncraft}   & $\checkmark$ & $\checkmark$ & $\checkmark$ & $\times$ & $\times$ & $\times$ & $\times$ & $\checkmark$  & 3   & 48.4 \\
      \midrule
      AR & MoMask~\cite{momask}   & $\checkmark$ & $\times$ & $\times$ & $\times$ & $\times$ & $\times$ & $\times$ & $\times$  & 2   & 28.6 \\
      AR & MotionGPT~\cite{motiongpt}   & $\checkmark$ & $\times$ & $\times$ & $\times$ & $\times$ & $\times$ & $\times$ & $\times$  & 2   & 28.6 \\
      AR & $M^{3}$GPT~\cite{m3gpt}   & $\checkmark$ & $\checkmark$ & $\times$ & $\times$ & $\times$ & $\times$ & $\checkmark$ & $\times$  & 3   & 164 \\
      \midrule
      AR-DiT & AMD~\cite{amd}   & $\checkmark$ & $\checkmark$ & $\times$ & $\times$ & $\times$ & $\checkmark$ & $\times$ & $\times$  & 4   & 85.87  \\
      AR-DiT & DART~\cite{dart}   & $\checkmark$ & $\times$ & $\times$ & $\checkmark$ & $\checkmark$ & $\checkmark$ & $\checkmark$ & $\times$  & 2   & 43.5 \\
      AR-DiT  & \OurMethod~(Ours)    & $\checkmark$ & $\checkmark$ & $\checkmark$ & $\checkmark$ & $\checkmark$ & $\checkmark$ & $\checkmark$ & $\checkmark$  & \textbf{28} & \textbf{286.2} \\ 
      \bottomrule
    \end{tabular}
  }
  \caption{Comparison between \OurMethod~and existing human motion generation methods. "GSTC(S)" and "GSTC(D)" denote global spatial-temporal controllable motion generation, where "S" and "D" indicate sparse and dense controlled joints, respectively. “Reference Motion” originates from user-designed or previously generated motion. "Mixed-condition" refers to the simultaneous occurrence of multiple conditions during training. “Datasets” indicates the total number of datasets used for training, while "Hours" represents the longest training dataset duration. For methods like MoMask, trained separately on HumanML3D or KIT, the duration of the HumanML3D dataset is considered.}
  \label{tab:method_compare}
\end{table*}

Multimodal whole-body human motion generation plays critical roles in animation~\cite{chen2024taming}, gaming~\cite{CombatMotion}, virtual reality~\cite{guo2024crowdmogen}, and embodied intelligence~\cite{mao2024learning} across diverse input conditions, including text, audio, and trajectory, etc.
The limited multimodal data and task-specific model designs prevent existing motion generation methods from supporting multimodal whole-body human motion generation. Due to the high cost of mocap data collection and labeling, most datasets focus on single domains such as Text-to-Motion (T2M)~\cite{humanml3d,kit}, Music-to-Dance (M2D)~\cite{finedance,aistpp}, Speech-to-Gesture (S2G)~\cite{liu2023emage,talkshow}, Human-Object Interaction (HOI)~\cite{behave, arctic, huang2022intercap}, Human-Scene Interaction (HSI)~\cite{trumans}, and Human-Human Interaction (HHI)~\cite{liang2024intergen, interx}, each with inconsistent data formats (\emph{e.g.}, BVH, SMPL-(H/X) and 3D Keypoints) and control conditions (\emph{e.g.}, Text captions, Audio, and Trajectories).
To tackle motion generation across diverse scenarios, it is crucial to build a unified framework that utilizes large-scale, diverse data to achieve more generalized representations.

Although recent works~\cite{mcm, zhang2024large, bian2024motioncraft,motionllama,amd} attempt to unify multitasks in one model, they meet challenges in multimodal control modeling, versatile tasks, and high-quality motion generation (as shown in Tab.~\ref{tab:method_compare}):
(1) \textbf{Independent Model Training.} Previous approaches train separate models for each modality, limiting simultaneous control across inputs~\cite{amd}.
(2) \textbf{Additional Control Branches.} Some methods add separate control branches for each condition, limiting interaction between them~\cite{bian2024motioncraft, mcm}.
(3) \textbf{Conflict Granularity Training.} Existing methods use mixed training by combining high-level semantic conditions with low-level controls, which hampers effective control at different levels and leads to optimization challenges~\cite{zhang2024large, m3gpt, motionllama}. Similar phenomena have also been observed in video generation~\cite{omnihuman,bodyofher}.
In addition to modeling challenges, relevant works also introduce large-scale motion datasets from multiple tasks and modalities~\cite{motionx,zhang2024large,liang2024omg,scamo,motionllama}, they still exhibit the following significant shortcomings (see comparisons in Tab.~\ref{tab:dataset_compare}):
(1) \textbf{Low Motion Quality.} Datasets expanded with non-mocap motion estimation exhibit ``garbage in, garbage out" effects, leading to poor-quality motion~\cite{motionx, scamo, zhang2024large}.
(2) \textbf{Text Inconsistency.} Inconsistent text annotations or expanded LLM descriptions lead to uneven text quality and hallucination issues~\cite{motionllama, scamo, liang2024omg}.
and (3) \textbf{limited tasks.} They focus on common tasks, limiting applicability in diverse ones like HOI, HSI, and HHI~\cite{zhang2024large, bian2024motioncraft, scamo}.
These limitations hinder the development of a unified, high-quality dataset for multimodal whole-body human motion generation across diverse scenarios.

\begin{table*}[!t]
  \centering
  \resizebox{\textwidth}{!}{ 
    \begin{tabular}{ccccccccccccc}
      \toprule
      \multirow{2}{*}{Dataset}  & \multicolumn{6}{c}{Tasks} & \multirow{2}{*}{Whole-Body} & \multirow{2}{*}{Motion Source}  & \multirow{2}{*}{Caption} & \multirow{2}{*}{Hierarchical} & \multirow{2}{*}{Frames} & \multirow{2}{*}{Hours} \\
      \cmidrule(lr){2-7}
       & T2M & M2D & S2G & HOI & HSI & HHI &  & Mocap / Total & Source & Caption &  &  \\
      \midrule
       Motion-X~\cite{motionx} &  $\checkmark$ & $\times$ & $\times$ & $\times$ & $\times$ & $\times$  & $\checkmark$  &  2 / 9 &   T (9)  & $\checkmark$ &   15.6M & 144.2  \\
       OMG~\cite{liang2024omg}  & $\checkmark$ & $\times$ & $\times$ & $\times$ & $\times$ & $\checkmark$  & $\times$   &  9 / 13  & \text{-}  & $\times$ &   22.3M & 206.5  \\
       MotionUnion~\cite{scamo}  & $\checkmark$ & $\times$ & $\times$ & $\times$ & $\times$ & $\times$  & $\times$   &  4 / 15 & \text{-} & $\times$ &   30M & 260  \\
       MotionVerse~\cite{zhang2024large}  & $\checkmark$ & $\checkmark$ & $\checkmark$ & $\times$ & $\times$ & $\times$  & $\checkmark$   & 5 / 16  & \text{-} & $\times$ &   \textbf{100M} & \text{-}  \\
      \midrule
      \TrainingData~(Ours)   & $\checkmark$ & $\checkmark$ & $\checkmark$ & $\checkmark$ & $\checkmark$ & $\checkmark$  & $\checkmark$   & \textbf{21} $/$ \textbf{28} & V + T (28) & $\checkmark$ &  64.3M & \textbf{286.2} \\
      \bottomrule
    \end{tabular}
  }
  \caption{Comparisons between \TrainingData~and existing merged datasets. "Mocap Source" indicates the proportion of mocap datasets. "Caption Source" specifies the method for completing missing descriptions: "\text{-}" (no completion), "V" (visual information), and "T" (textual information). "Hierarchical Caption" shows if captions include hierarchical text.
  }
  \vspace{-7pt}
  \label{tab:dataset_compare}
\end{table*}

To address existing challenges, we propose \OurMethod, a unified framework for multimodal whole-body human motion generation, and \TrainingData, the largest unified multimodal mocap motion dataset. 
Specifically, \OurMethod~employs Diffusion Transformers (DiTs)~\cite{DIT}, incorporates multimodal conditions by concatenating condition tokens as prefix context, and adopts a progressive weak-to-strong mixed conditions training strategy to gradually constrain motion from high-level semantics to dense spatial-temporal alignment. 
Notably, unlike previous methods, \OurMethod~introduces a novel generation paradigm that utilizes reference motion (\emph{e.g.}, user-provided or model-predicted motion) as a special condition. This significantly enhances generated motion quality and achieves consistency between reference and generated motion, creating an effective clip-by-clip autoregressive motion diffusion.
This enables \OurMethod~to support autoregressive interactive generation with strong temporal alignment.
Furthermore, \OurMethod~unifies various Global Spatial-Temporal Controllable Generation tasks through spatial-temporal masking strategies. 
To guarantee high-quality motion generation training, we collect high-quality mocap datasets that support diverse motion generation tasks, unify them under the SMPL-X~\cite{smplx} format with standard world coordinate systems, and automatically generate hierarchical text captions by rendering motions into videos and annotating them with vision language models (VLMs). 
This dataset contains about \emph{$286.2$ hours} and integrates multimodal control conditions, supporting versatile tasks, including T2M, M2D, S2G, HOI, HSI, and HHI.

In summary, our core contributions are as follows:
\begin{itemize}
    \item We propose \OurMethod, a multimodal autoregressive diffusion transformer for versatile whole-body motion generation. By introducing reference motion, \OurMethod~significantly enhances consistent content, style, and temporal dynamics generation.
    
    \item We propose a progressive weak-to-strong mixed conditions training strategy to effectively handle multi-granular constraints.
    
    \item We construct \TrainingData, the largest unified multimodal mocap motion dataset with a unified SMPL-X format, and provides consistent, detailed, and structured text captions to serve versatile motion generation tasks.
    
    \item Extensive experiments show that \OurMethod~achieves state-of-the-art performance across various tasks, including T2M, S2G, M2D, and GSTC. These evaluations were conducted on our more challenging test sets comprising $280$ samples uniformly sampled from our \TrainingData~across diverse scenarios.
\end{itemize}

\section{Related Work}
\label{sec:formatting}

\subsection{Human Motion Generation}

Existing methods are typically categorized based on input conditions into three main types: single-modal, cross-modal, and multimodal. 
Single-modal motion generation uses control conditions from the same modality as the motion, including motion prediction~\cite{xu2024learning,humanmac}, in-betweening~\cite{cohan2024flexible}, joint/trajectory-guided synthesis~\cite{omnicontrol,motionlcm,dart}, and body-shape-conditioned motion generation~\cite{tripathi2024humos, xue2025shape}. 
Cross-modal motion generation uses control conditions from different modalities, including text in T2M~\cite{mdm,mld,momask,humantomato,motiongpt,motiongptv2, liang2024omg,t2mgpt,motiondiffuse}, music in M2D~\cite{finedance,aioz,bailando,edge}, speech in S2G~\cite{liu2023emage,talkshow,ted-smplx,chen2024diffsheg}, target object positions in HOI~\cite{arctic,taco,yang2025smgdiff}, scene layouts in HSI~\cite{emdb,rich,trumans}, and partner movements in HHI~\cite{interx,liang2024intergen,wang2025timotion}.
However, these approaches typically rely on task-specific architectures, significantly limiting their cross-task generalization capabilities and practical applications.
Recently, researchers~\cite{amd,mcm,li2025genmo,zhang2024large,bian2024motioncraft,m3gpt} have begun exploring multimodal motion generation, with approaches falling into three primary categories:

\begin{itemize}
    \item \textbf{Separate Model Training.} These approaches~\cite{amd} typically train independent models for different conditional modalities, making multimodal motion generation fundamentally unattainable.
    
    \item \textbf{Additional Control Branch.} These approaches~\cite{bian2024motioncraft,mcm} incorporate separate control branches into the backbone architecture, each dedicated to a specific condition, resulting in limited interaction between different conditions.
    
    \item \textbf{Unified Multimodal Condition Modeling.} These approaches~\cite{zhang2024large,motionllama,yang2025unimumo,zhang2025motion} use a single model to learn mappings from diverse modalities to target motions, enhancing multi-modal adaptability. However, challenges arise from different constraints across modalities: text provides semantic guidance, spatial-temporal controls impose physical constraints, while audio inputs enforce rhythmic alignment. These disparate constraints often result in compromised controllability and optimization difficulties.
    
\end{itemize}

\subsection{Human Motion Dataset}
Current human motion generation datasets are predominantly task-specific. 
Different tasks rely on separate collections: text-to-motion (T2M) uses datasets with text captions~\cite{humanml3d, kit, motionx, babel, CombatMotion}, music-to-dance (M2D) requires music-annotated datasets~\cite{aistpp, finedance, aioz, choreomaster}, and speech-to-gesture (S2G) employs datasets with paired speech~\cite{liu2023emage, talkshow, liu2022beat, ted-smplx}. 
Similarly, interaction datasets remain disconnected across human-object~\cite{behave, arctic, jiang2022chairs, li2023object, hoim3, taco, oakink2}, human-scene~\cite{trumans}, and human-human interaction~\cite{liang2024intergen, interx}. This fragmentation constrains multimodal motion generation capabilities.
Integrating these datasets presents challenges due to inconsistent motion formats. 
Datasets vary widely in their representations: some use SMPL~\cite{behave, liang2024intergen, aistpp, fan2025go}, others employ SMPL-X~\cite{liu2023emage, talkshow, ted-smplx}, while many rely on BVH~\cite{lafan, 100style, mixamo, motorica_1,motorica_2} or keypoint representations~\cite{human3.6m, uestc}. This diversity makes format standardization extremely difficult. 
Recent works~\cite{motionx, liang2024omg, scamo, zhang2024large, motionllama} have attempted to integrate multiple datasets, increasing scale and incorporating some multimodal conditions. However, they still exhibit several limitations:

\begin{itemize}
    \item \textbf{Low-quality Motions}: Most integrated datasets derive from non-mocap sources with significant estimation errors~\cite{motionx, scamo, zhang2024large, motionllama, fan2025go}, creating a ``garbage in, garbage out" effect that compromises motion generation quality.

    \item \textbf{Text Inconsistency}: Existing datasets~\cite{liang2024omg, zhang2024large, scamo} use inconsistent text annotations (\emph{e.g.}, action labels versus semantic descriptions) or rely on LLMs for text refinement without visual motion data, causing variable text quality and hallucinations~\cite{motionx, motionllama}.
    
    \item \textbf{Limited Generation Scenarios}: Most datasets lack support for critical interaction tasks including HOI, HSI~\cite{motionx, liang2024omg, scamo, zhang2024large, motionllama}, and HHI~\cite{motionx, scamo, zhang2024large}, severely limiting their applicability to complex scenarios.
\end{itemize}

\section{OmniMoCap-X Dataset}

To address the data challenges—the scarcity of high-quality multitask motion datasets, inconsistent textual annotations and motion representations—we implement a three-pronged approach: (1) integrating diverse motion capture datasets with multitask support, (2) establishing a unified motion representation, and (3) developing high-quality motion captions derived from visual-textual annotations.

\begin{table}[h]
  \centering
  \setlength{\tabcolsep}{1pt} 
  {\small 
  \begin{tabular}{@{}clcccc@{}}
    \toprule
    Task & Dataset & Frames & Hours &  Mocap & Format \\
    \midrule
    \multirow{6}{*}{\makecell{T2M}} & Mixamo~\cite{mixamo} & 0.4M & 1.9 & Marker-M & BVH\\
    & KIT~\cite{kit} & 2.3M & 6.4 & Marker-V & SMPL \\
    & OMOMO~\cite{li2023object} & 1.1M & 2.5 & Marker-V & SMPL-X\\
    & IDEA400~\cite{motionx} & 1.7M & 15.7 & SV-RGB & SMPL-X\\
    & 100Style~\cite{100style} & 4.8M & 22.1 & IMU & BVH \\
    & HumanML3D~\cite{humanml3d} & 26.5M & 65.7 & Marker-V & SMPL\\
    \midrule
    \multirow{6}{*}{\makecell{M2D}} & Choreomaster~\cite{choreomaster} & 0.1M & 1.2 & Marker-M & FBX\\
    & Finedance~\cite{finedance} & 0.8M & 7.7 & Marker-V & SMPL-H \\
    & Phantomdance~\cite{li2022danceformer} & 1.0M & 9.5 & Marker-M & SMPL \\
    & AIST++~\cite{aistpp} & 1.1M & 5.2 & MV-RGB & SMPL \\
    & Motorica~\cite{motorica_1} & 2.7M & 12.4 & Marker-V & BVH\\
    & AIOZ~\cite{aioz} & 6.6M & 60.8 & SV-RGB & SMPL \\
    \midrule

    \multirow{1}{*}{\makecell{S2G}} & BEAT2~\cite{liu2023emage} & 6.9M & 64.2 & Marker-V & SMPL-X\\
    \midrule
    \multirow{3}{*}{\makecell{HHI}} 
    & Humansc3d~\cite{humansc3d} & 0.2M & 1.1 & Marker-V & SMPL-X \\
    & InterHuman~\cite{liang2024intergen} & 4.5M & 20.8 & MV-RGB & SMPL\\
    & Inter-x~\cite{interx} & 16.2M & 37.5 & Marker-V & SMPL-X \\
    \midrule
    \multirow{8}{*}{\makecell{HOI}} & Arctic~\cite{arctic} & 0.2M & 2.0 & Marker-V & SMPL-X \\
    & TACO~\cite{taco} & 0.4M & 3.3 & Marker-V & MANO \\
    & Fit3d~\cite{fit3d} & 0.4M & 2.5 & Marker-V & SMPL-X \\
    & Behave~\cite{behave} & 0.4M & 4.1 & MV-RGB & SMPL-X \\
    & Chairs~\cite{jiang2022chairs} & 1.0M & 9.5 & Marker-V & SMPL-X\\
    & HOI-M3~\cite{hoim3} & 2.3M & 10.7 & MV-RGB & SMPL \\
    & Oaklnkv2~\cite{oakink2} & 4.0M & 36.8 & Marker-V & SMPL-X \\
    \midrule
    \multirow{5}{*}{\makecell{HSI}} & EMDB~\cite{emdb} & 0.03M & 0.3 & IMU & SMPL \\
    & Rich~\cite{rich} & 0.1M & 0.8 & MV-RGB & SMPL-X \\
    & Lafan1~\cite{lafan} & 1.0M & 4.5 & Marker-V & BVH \\
    & Trumans~\cite{trumans} & 1.2M & 11.2 & Marker-V & SMPL-X \\
    & Circle~\cite{circle} & 4.4M & 10.1 & Marker-V & SMPL-X \\
    \midrule
    All & \TrainingData & 64.3M & 286.2 & Mixed & SMPL-X \\
    \bottomrule
  \end{tabular}
  } 
  \caption{
  Composition of \TrainingData~dataset, unifying motion formats into SMPL-X with captions. We select $28$ publicly available high-quality datasets across various tasks. Frames and Hours are computed based on raw dataset FPS. MoCap represents data capture methods, ranked by quality: Marker with manual correction (Marker-M), Vicon Marker (Marker-V), IMU, Multi-View RGB (MV-RGB), and Single-View RGB (SV-RGB). Format specifies the original motion format.}
  \label{tab:dataset_information}
\end{table}

\subsection{Mocap Dataset Composition}
While existing approaches prioritize dataset scale over quality by incorporating substantial amounts of non-mocap data~\cite{motionx, scamo, zhang2024large, motionllama}, leading to degraded motion generation quality ("garbage in, garbage out"), our work emphasizes data quality.
We systematically curate open-source mocap datasets and classify them into five acquisition categories: Marker with manual correction, Marker (Vicon), IMU, Multi-View RGB, and Single-View RGB (detailed in Tab.~\ref{tab:dataset_information}).
Our dataset supports diverse motion generation tasks, including text-to-motion, motion-to-dance, speech-to-gesture, and various interactions (human-object, human-scene, and human-human), providing a high-quality multimodal foundation. Comprising $64.3$ million frames and $286.2$ hours of data, our approach achieves comprehensive task coverage while maintaining superior data quality. 
Visualizations of specific interaction scenarios are showcased in the supplementary material.

\subsection{Unified Motion Representation}
We standardize diverse motion formats (BVH~\cite{bvh}, FBX~\cite{wikipediafbx}, and SMPL~(-H)~\cite{smpl}) into the SMPL-X~\cite{smplx} format to support our unified multimodal model.
First, we convert these formats into the Motion-X~\cite{motionx} SMPL-X format, comprising root orientation, pose parameters (body, hand, and jaw), facial expressions, facial shape, translation, and body shape parameters.
We then normalize the translation scale and initial root orientation across datasets to establish a consistent coordinate system. Additionally, we resample all datasets to 30 frames per second (FPS) to enhance the model's ability to capture temporal patterns. 
The specific conversion process and normalization for different datasets are provided in the supplementary material.

To enhance generalization ability, we extend the widely-used body-only representation~\cite{humanml3d} to a whole-body format. 
Specifically, the pose of the $i$-th frame is a tuple $\mathbf{p}_i = ({\dot{r}^a, \dot{r}^x, \dot{r}^z, r^y, \mathbf{j}^p, \mathbf{j}^v, \mathbf{c}^f, \mathbf{f}})$, where $\dot{r}^a\in \mathbb{R}$ is root angular velocity around the Y-axis; $\dot{r}^x, \dot{r}^z \in \mathbb{R}$ are root linear velocity in the XZ plane; $r^y \in \mathbb{R}$ is root height, $\mathbf{j}^p \in \mathbb{R}^{3N-1}$ are local joint positions; Notably, $\mathbf{j}^r \in \mathbb{R}^{6N'}$ are joint 6D rotations of SMPL-X; $\mathbf{j}^v \in \mathbb{R}^{3N}$ are joint velocities; $\mathbf{c}^f$ is binary features obtained by thresholding the heel and toe joint velocities to emphasize the foot ground contacts. 
Here, $N$ refers to 127 whole-body joints extracted from SMPL-X~\cite{smplx}, while $N'$ represents 53 joints spanning the body, hands, and jaw.
For facial representation, we adopt the Flame Format~\cite{flame100}, encoding facial features as $\mathbf{f} \in \mathbb{R}^{100}$.

\begin{figure*}[t]
  \centering
  \includegraphics[width=\textwidth]{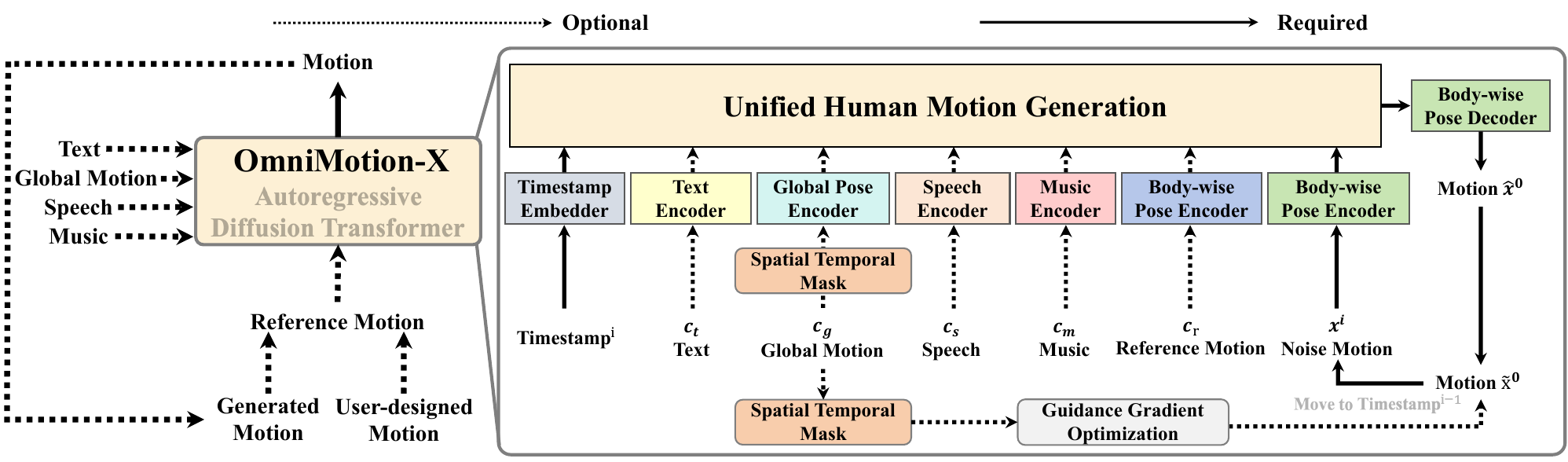}
  \caption{
  Overview of \OurMethod, a unified multimodal autoregressive transformer diffusion model for whole-body human motion generation. \OurMethod~integrates text, global motion, speech, music, and reference motion as conditions through condition-specific encoders mapped into a unified space. The model fuses multimodal information to produce coherent motion, with spatial-temporal guidance ensuring consistent global motion characteristics.
  }
  \vspace{-5pt}
  \label{fig:method}
\end{figure*}

\subsection{Consistent Visual-textual Motion Captions}

Current datasets employ three annotation approaches:
\begin{itemize}
    \item \textbf{LLM Refinement}~\cite{motionx, scamo} employs LLMs like GPT-4~\cite{achiam2023gpt} to enhance textual descriptions. However, since LLMs cannot directly perceive motion data and rely solely on existing text, they frequently introduce hallucinations and fail to capture precise action details.
    
    \item \textbf{Manual Annotation}~\cite{humanml3d, babel, motionllama} effectively eliminates hallucinations but proves costly and difficult to scale, resulting in limited annotations with simple descriptions.
    \item \textbf{Motion-to-Text Model-Based Annotation}~\cite{posescript,motionscript} generates captions directly from raw motion. While these methods effectively capture low-level kinematic details (\emph{e.g.}, left elbow bends 45 degrees), they lack the understanding of high-level semantics and action context.
\end{itemize}

To overcome these limitations, we propose an automated approach that integrates both visual and textual information for comprehensive motion annotation.
Our method renders motion into videos and combines them with existing textual annotations (\emph{e.g.}, descriptions, action labels, and task categories) as input to the state-of-the-art vision-language models (VLMs), GPT-4o~\cite{achiam2023gpt}.
This multimodal approach generates structured, hierarchical, and precise motion captions that ensure both annotation quality and expression richness. The quality of the texts was ensured through iterative prompt optimization and a comparative evaluation of various VLMs, with specific examples and detailed caption statistics provided in the supplementary material.

\begin{table*}
  \centering
  \resizebox{\textwidth}{!}{
    \begin{tabular}{cccccccc}
      \toprule
        \multicolumn{1}{c}{\multirow{2}{*}{Method}} & \multicolumn{3}{c}{R Precision$\uparrow$} & \multicolumn{1}{c}{\multirow{2}{*}{FID$\downarrow$}} & \multirow{2}{*}{Multimodal Dist.$\downarrow$} & \multicolumn{1}{c}{\multirow{2}{*}{Diversity$\rightarrow$}} & \multicolumn{1}{c}{\multirow{2}{*}{MultiModality$\uparrow$}} \\
        \cmidrule(lr){2-4} 
        & Top-1 & Top-2 & Top-3 & & & & \\
      \midrule
      GT & $0.535^{\pm0.009}$ & $0.725^{\pm0.009}$ & $0.821^{\pm0.008}$ & $0.013^{\pm0.005}$ & $2.493^{\pm0.011}$ & $9.194^{\pm0.093}$ & - \\
      \midrule
      MDM~\cite{mdm} & $0.063^{\pm0.004}$ & $0.121^{\pm0.003}$ & $0.169^{\pm0.003}$ & $72.928^{\pm0.361}$ & $8.376^{\pm0.018}$ & $1.981^{\pm0.003}$ & $0.394^{\pm0.010}$ \\
      MLD~\cite{mld} & $0.084^{\pm0.002}$ & $0.152^{\pm0.002}$ & $0.209^{\pm0.003}$ & $70.082^{\pm0.468}$ & $8.281^{\pm0.025}$ & $1.998^{\pm0.003}$ & ${0.439}^{\pm0.022}$ \\
      MoMask~\cite{momask} & $0.104^{\pm0.003}$ & $0.163^{\pm0.003}$ & $0.199^{\pm0.005}$ & $69.361^{\pm0.365}$ & $8.341^{\pm0.022}$ & $2.123^{\pm0.002}$ & $0.482^{\pm0.008}$ \\
      MoMask*~\cite{momask} & $0.267^{\pm0.004}$ & ${0.414}^{\pm0.004}$ & ${0.530}^{\pm0.004}$ & ${
      17.428}^{\pm0.400}$ & ${5.661}^{\pm0.024}$ & ${6.772}^{\pm0.013}$ & $0.811^{\pm0.067}$ \\
      MotionCraft~\cite{bian2024motioncraft} & ${0.176}^{\pm0.004}$ & $0.259^{\pm0.003}$ & $0.319^{\pm0.004}$ & $63.049^{\pm0.470}$ & $7.936^{\pm0.021}$ & $2.325^{\pm0.013}$ & $0.557^{\pm0.039}$ \\
      MotionCraft*~\cite{bian2024motioncraft} & $0.236^{\pm0.004}$ & ${0.370}^{\pm0.004}$ & ${0.489}^{\pm0.004}$ & ${47.428}^{\pm0.400}$ & ${7.424}^{\pm0.024}$ & ${2.820}^{\pm0.013}$ & $0.863^{\pm0.067}$ \\
      \midrule
      \OurMethod~(Ours) & \underline{$0.303^{\pm0.004}$} & \underline{$0.464^{\pm0.006}$} & \underline{$0.571^{\pm0.005}$} & \underline{$5.040^{\pm0.293}$} & \underline{$4.678^{\pm0.019}$} & $\mathbf{8.650}^{\pm0.095}$ & $\mathbf{1.696}^{\pm0.366}$ \\
      \OurMethod~(Ours + RM) & $\mathbf{0.346}^{\pm0.005}$ & $\mathbf{0.511}^{\pm0.007}$ & $\mathbf{0.629}^{\pm0.006}$ & $\mathbf{3.199}^{\pm0.293}$ & $\mathbf{4.106}^{\pm0.019}$ & $8.009^{\pm0.095}$ & $1.143^{\pm0.366}$ \\
      \bottomrule
    \end{tabular}}
  \caption{Quantitative results of text-to-motion on the \TrainingData~test set. $\uparrow$ ($\downarrow$) indicates that a larger (smaller) value is better. $\rightarrow$ indicates that a value closer to the GT is better. Bold entries indicate the best results, and underlined entries indicate the second-best results. All evaluations are repeated 20 times, reporting the mean and 95\% confidence interval.}
  \vspace{-5pt}
  \label{tab:T2M}
\end{table*}
\section{Method}

The overview of \OurMethod~is illustrated in Fig.~\ref{fig:method}.
We propose a unified multimodal, autoregressive transformer-based diffusion model for whole-body human motion generation across multiple tasks.
The model takes textual description \(\mathbf{c}_{t}\) for semantic guidance, global motion \(\mathbf{c}_{g}\) for spatial-temporal control, and speech \(\mathbf{c}_{s}\) and music conditions \(\mathbf{c}_{m}\) to ensure rhythmic and stylistic coherence. 
Furthermore, it incorporates reference motion \(\mathbf{c}_{r}\) as a motion prior, derived from either previously generated clips or user-designed motion, thus providing fine-grained details unavailable in other conditions. 
This reference input is optional and can be set to null, which is essential for generating the initial motion when no user-specified prior exists.
In addition, the mixed-condition setting allows multiple conditions to occur simultaneously during training, enabling the model to handle complex scenarios with diverse inputs.
In this section, we present our approach in two parts: a unified framework for multimodal modeling and a progressive weak-to-strong mixed conditions training strategy that ensures precise motion control.

\begin{figure}
    \centering
    \includegraphics[width=\columnwidth]{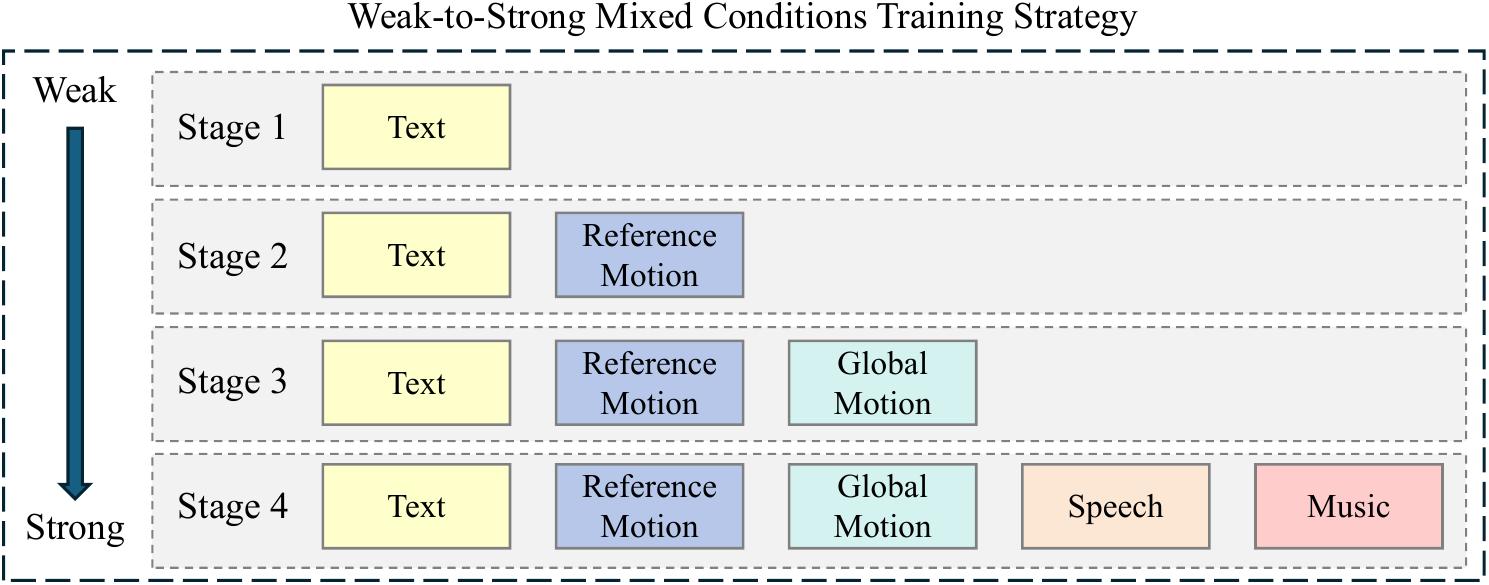}
    \caption{We propose the weak-to-strong conditions training strategy, establishing motion-semantic alignment with text, followed by progressive integration of stronger multimodal signals (reference motion, global motion, speech, music) for enhanced generation quality and controllability.}
    \label{fig:training_strategy}
    \vspace{-1pt}
\end{figure}


\subsection{Unified Multimodal Modeling}
\label{sec:Unified_MultiModal}

We integrate multimodal conditions using Diffusion Transformers (DiTs)~\cite{DIT} by concatenating condition tokens as prefix contexts, allowing the model to process and fuse multimodal information efficiently.

\textbf{Multimodal Conditions.} Our framework integrates multiple condition modalities: text \(\mathbf{c}_t\) providing semantic guidance; global motion \(\mathbf{c}_g\) ensuring spatial-temporal consistency; speech \(\mathbf{c}_s\) synchronizing gestures and lip movements with rhythm; music \(\mathbf{c}_m\) supplying beat and style information for dance; and reference motion \(\mathbf{c}_r\) serving as a motion prior. 
%
Notably, this reference motion condition—overlooked in previous multimodal motion generation~\cite{zhang2024large, mcm, bian2024motioncraft}—enables our model to maintain precise spatial-temporal pattern consistency with the reference, substantially enhancing motion quality and coherence. Crucially, a reference motion provides rich spatio-temporal context that a single starting pose cannot. By typically spanning a duration similar to the target sequence, it captures fine-grained details and style, which are absent in a momentary posture.
To fully leverage each modality and facilitate interaction between different conditions, we employ modality-specific encoders (\emph{e.g.}, T5-XXL~\cite{t5} for text, a wav encoder~\cite{liu2023emage} for speech, Librosa~\cite{librosa} for music and body-wise encoding~\cite{bian2024motioncraft} for motion) to extract features from each modality. These features are aligned to match motion embedding dimensions using learnable linear projections, allowing us to concatenate all condition tokens as prefix context with noisy motion tokens during processing.

\textbf{Multimodal Modeling.} 
Overall, the unified multimodal modeling is formulated as follows:  
\begin{equation}
\scalebox{0.78}{$c = [h_t ( f_t (c_t)), h_g (f_g (c_g)), h_s ( f_s (c_s)), h_m ( f_m (c_m)), h_r ( f_r (c_r))],$}
\label{eq:c_representation}
\end{equation}
where $f$ and $h$ denote the modality-specific encoder and projection layer, respectively. The concatenated representation $c$ is then fed into our DiT backbone as a conditioning prefix context, where attention mechanisms are employed to learn the correspondences among the various modalities.
To constrain the physical properties of motion, we follow previous works~\cite{mdm,chen2024taming} by directly predicting motion $\hat{x}_0$ rather than noise. Consequently, our diffusion objective is defined as follows:
\begin{equation}
    \mathcal{L}_{\text{simple}} = \mathbb{E}_{x_0 \sim q(x_0 | c), t \sim [1,T]} 
    \left[ \left\| x_0 - G(x_t, t, c) \right\|_2^2 \right],
    \label{eq:simple_loss}
\end{equation}
where $q(x_0 | c)$ represents the data distribution, $T$ is the maximum diffusion step. $G$ denotes the learned denoising function, while $x_t$ represents the noisy motion at step $t$, expressed as $x_t = [p_0^t, p_1^t, \dots, \mathbf{p}_N^t]$, where each $\mathbf{p}_i^t$ corresponds to the $i_{th}$ pose in motion at step $t$.
Further details are available in the supplementary material.

\subsection{Weak-to-Strong Progressive Training Strategy}
\label{sec:training_strategy}
Considering our diverse multi-granularity conditioning inputs, we empirically observed that employing all conditions within a single-stage training paradigm leads to difficulties in directly learning the correlation between motion and conditions. Moreover, the model tends to overfit strongly constrained low-level control signals such as fine-grained reference motion and spatiotemporal joint control. While these signals appear dominant, they can suppress other modalities such as text, ultimately compromising overall controllability.
To address this, we implement weak-to-strong mixed conditions training strategy: as shown in Fig.~\ref{fig:training_strategy} initial text-conditioned learning establishes motion-semantic alignment, followed by progressive integration of finer-grained conditions, including reference motion, global motion, and audio signals.
This progressive approach enables the model to effectively adapt to different conditions, ensuring high-quality and flexible motion generation while precisely adhering to multimodal conditions.

\begin{figure*}[h]
  \centering
  \subfloat[Text-to-motion]{\includegraphics[width=0.33\textwidth]{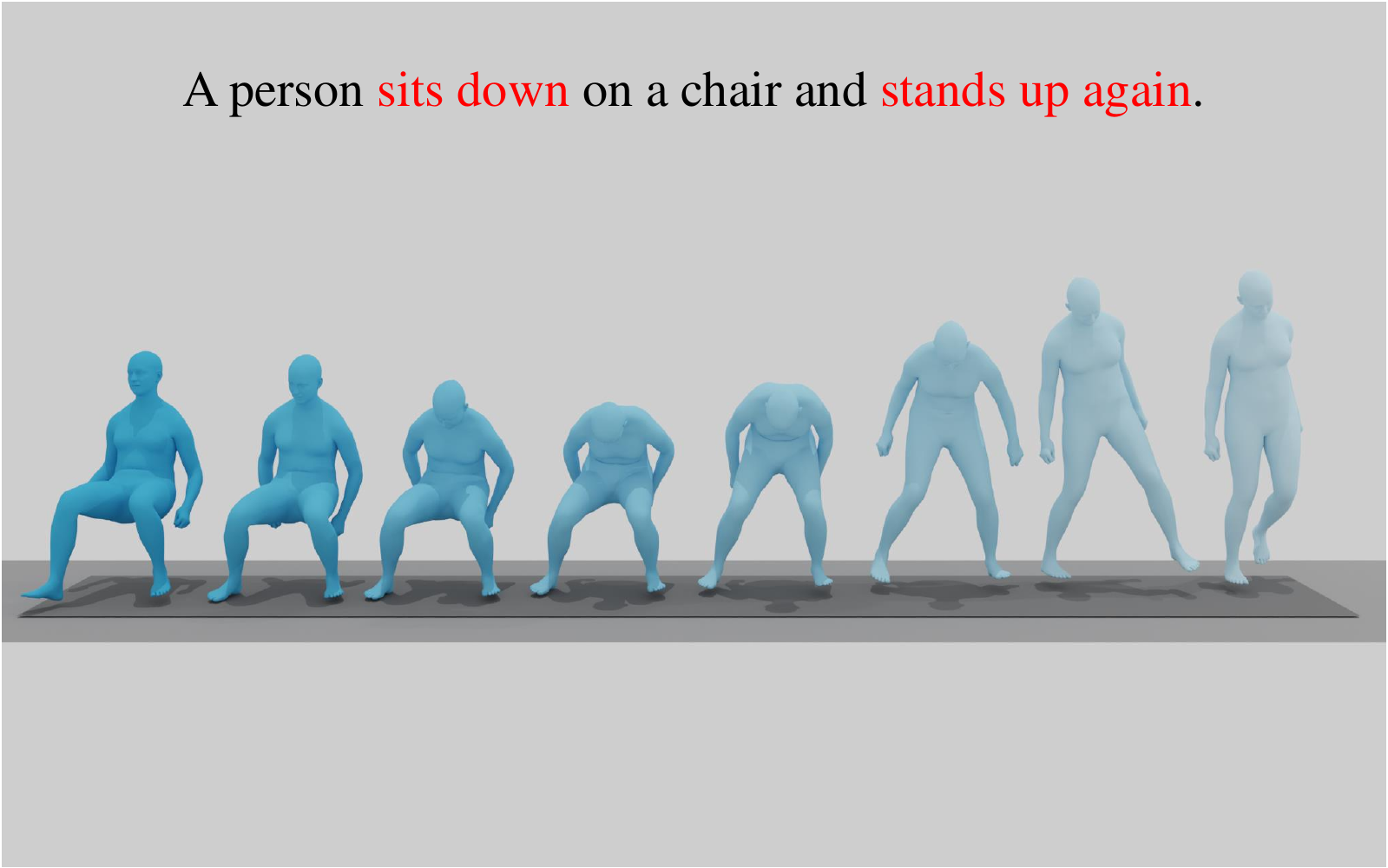}} \hfill
  \subfloat[Speech-to-gesture]{\includegraphics[width=0.33\textwidth]{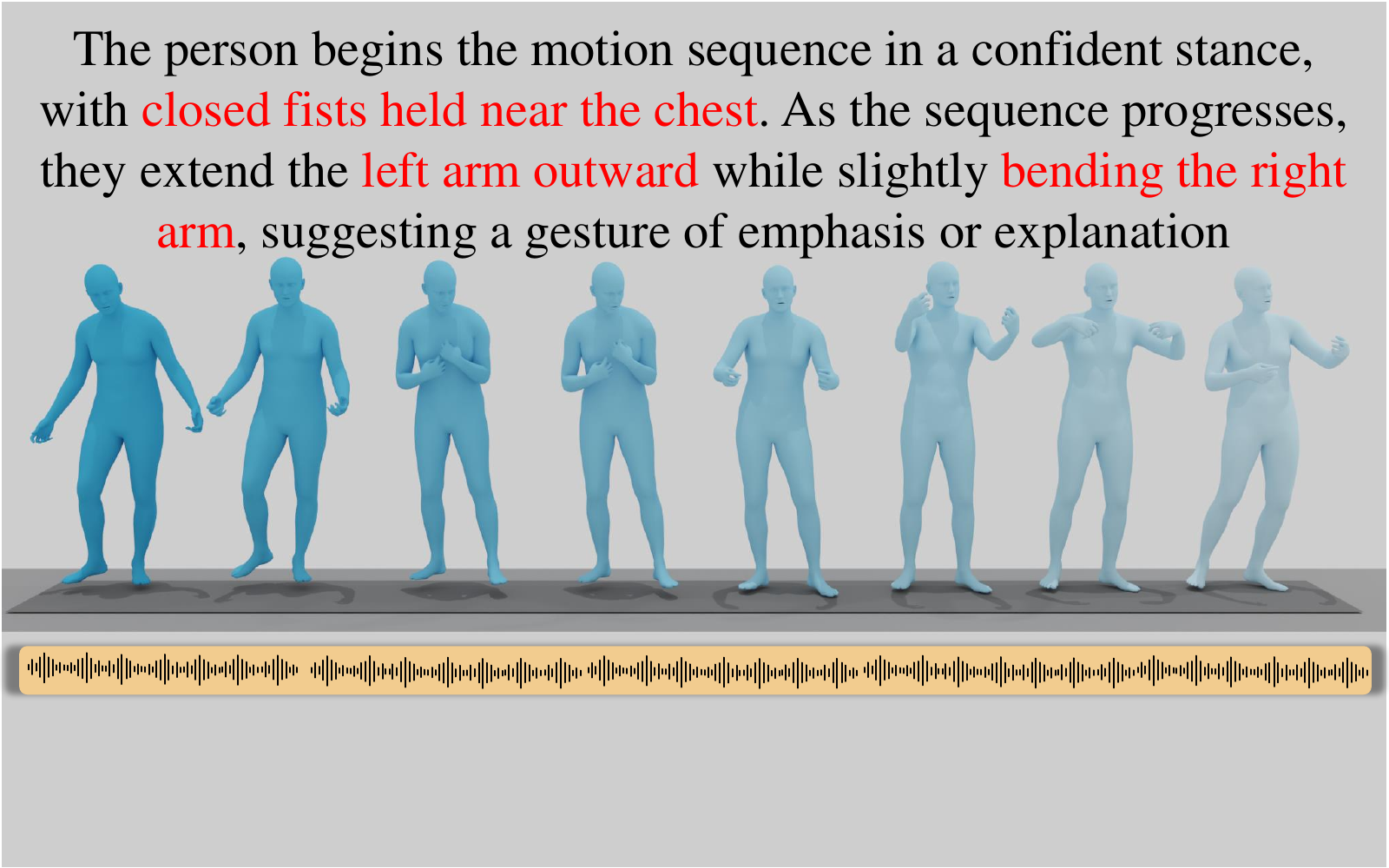}} \hfill
  \subfloat[Music-to-dance]{\includegraphics[width=0.33\textwidth]{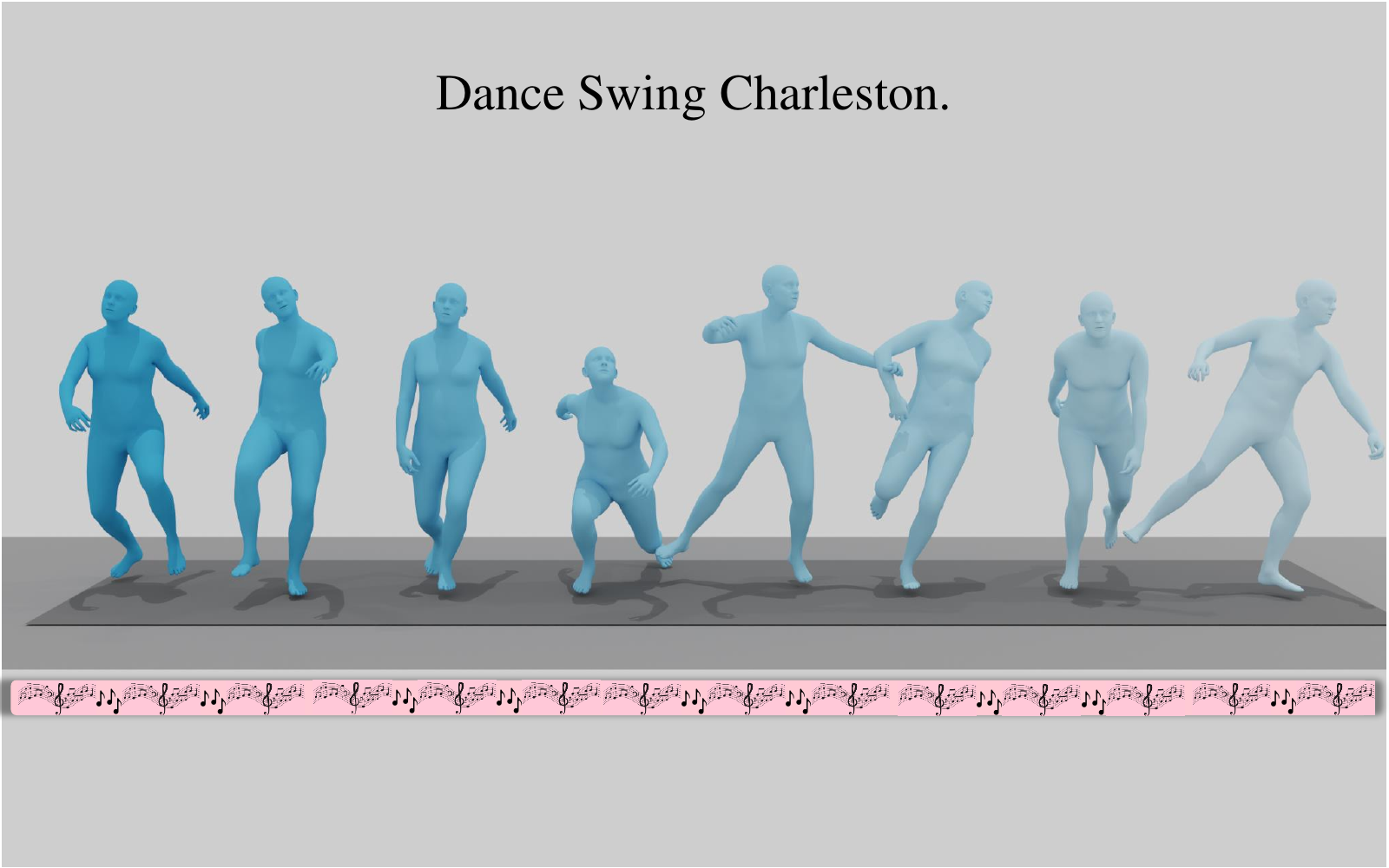}} \\
  \subfloat[Trajectory-guided synthesis]{\includegraphics[width=0.33\textwidth]{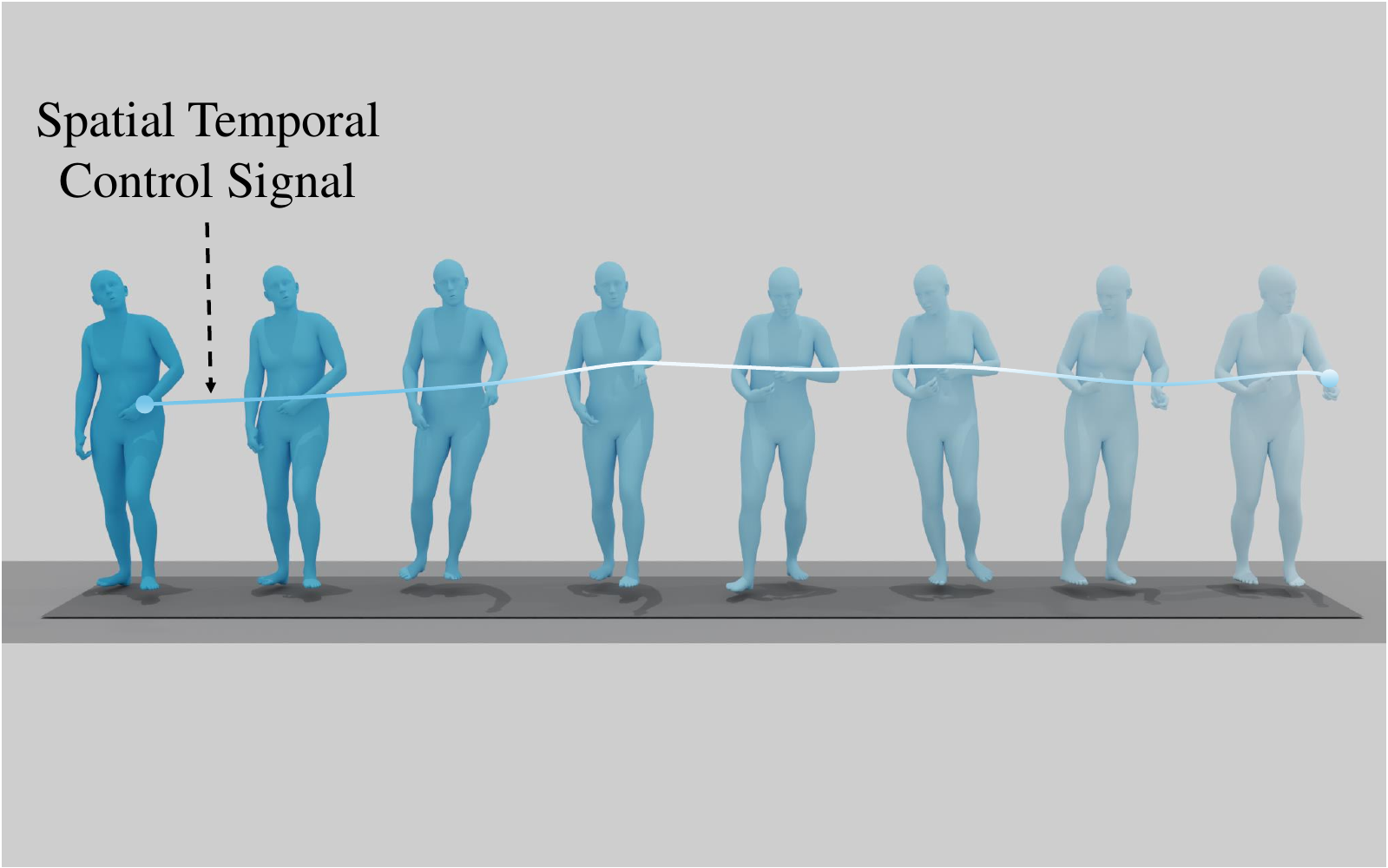}} \hfill
  \subfloat[Motion In-bewteen]{\includegraphics[width=0.33\textwidth]{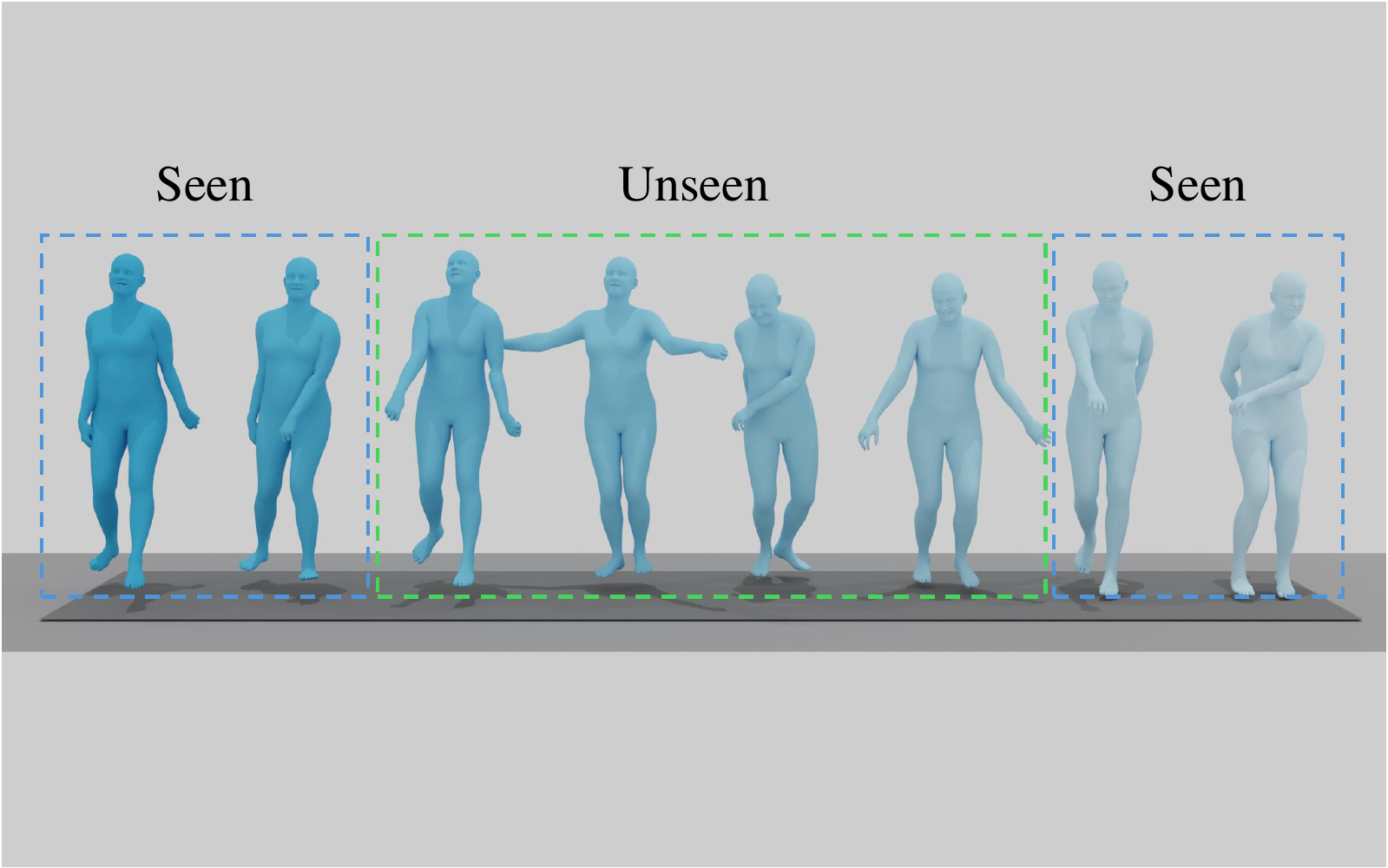}} \hfill
  \subfloat[Motion Prediction]{\includegraphics[width=0.33\textwidth]{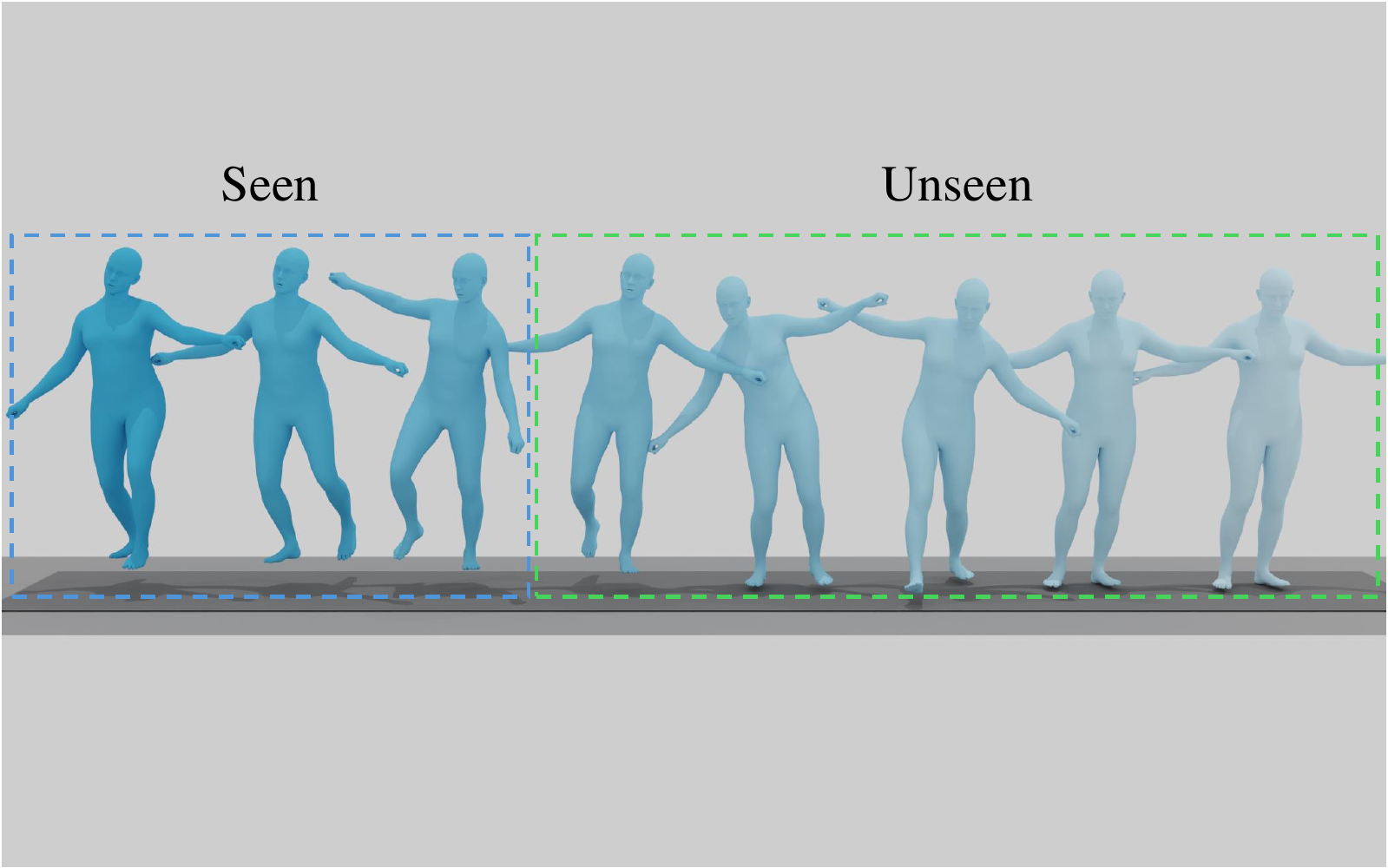}} \\
  \caption{\textbf{Diverse motion synthesis capabilities of \OurMethod.} 
  \OurMethod~supports multiple tasks: (a) text-to-motion, (b) speech-to-gesture, (c) music-to-dance, (d) trajectory-guided motion, (e) motion in-betweening, and (f) motion prediction.
  }
  \label{fig:Diverse motion synthesis capabilities}
\end{figure*}

\section{Experiments}
\begin{table}
  \centering
  \resizebox{\columnwidth}{!}{
    \begin{tabular}{lccccccc}
      \toprule
        \multirow{2}{*}{\textbf{Method}} & \multicolumn{3}{c}{\textbf{Text-motion quality}} & \multicolumn{4}{c}{\textbf{Geometric control}} \\
        \cmidrule(lr){2-4} \cmidrule(lr){5-8}
        & \textbf{FID$\downarrow$} & \textbf{R Prec.$\uparrow$} & \textbf{Multi.$\downarrow$} & \textbf{Ctrl L2$\downarrow$} & \textbf{Skating$\downarrow$} & \textbf{Fail@20$\downarrow$} & \textbf{Fail@50$\downarrow$} \\
      \midrule
      GT & 0.013 & 0.821 & 2.493 & - & - & - & - \\
      \midrule
      OmniControl & \underline{63.725} & \underline{0.392} & \underline{8.011} & \underline{0.820} & \underline{0.089} & \underline{0.844} & \underline{0.794} \\
      \OurMethod & \textbf{4.224} & \textbf{0.682} & \textbf{4.377} & \textbf{0.424} & \textbf{0.004} & \textbf{0.516} & \textbf{0.330} \\
      \bottomrule
    \end{tabular}
  }
  \caption{Quantitative results of global spatiotemporal controllable generation on the \TrainingData~test set. We report text-motion quality metrics and geometric control metrics in a unified table. Bold entries indicate the best results, and underlined entries indicate
the second-best results.}
  \vspace{-10pt}
  \label{tab:gstc}
\end{table}

\begin{table*}
  \centering
  \resizebox{\textwidth}{!}{ 
    \begin{tabular}{cccccccc}
      \toprule
      \multirow{2}{*}{Method} & \multicolumn{4}{c}{Speech-to-Gesture} & \multicolumn{3}{c}{Music-to-Dance} \\
      \cmidrule(lr){2-5} \cmidrule(lr){6-8}
      & FID (Whole-Body)$\downarrow$ & FID (Hands)$\downarrow$& $Face_{MSE}\downarrow$ & Diversity$\uparrow$ & FID (Whole Body)$\downarrow$ & FID (Hands)$\downarrow$ & Diversity$\uparrow$ \\
      \midrule
      MotionCraft & \underline{3.422} & \textbf{5.370} & \underline{0.182} & \underline{1.003} & \textbf{9.875} & \underline{7.099} & \underline{3.798}  \\
      \OurMethod (Ours) & \textbf{2.641} & \underline{9.095} & \textbf{0.045} & \textbf{1.664} & \underline{16.209} & \textbf{5.827} & \textbf{4.716}  \\
      \bottomrule
    \end{tabular}
  }
  \caption{\textbf{Results of S2G in BEAT2 and M2D in AIST++, FineDance and Phantomdance.} We respectively evaluate the $FID_{Whole Body}$,  $FID_{Hands}$, $Face_{MSE}$, and diversity for S2G and the $FID_{Whole Body}$,  $FID_{Hands}$, and diversity for M2D. Bold entries indicate the best results, and underlined entries indicate the second-best results.}
  \vspace{-8pt}
  \label{tab:s2g_m2d}
\end{table*}

\subsection{Implementation Details}
Our method employs a Transformer Encoder architecture~\cite{attention} with 8 layers and 8 attention heads, featuring a hidden dimension of \( d_{\text{model}} = 1536~(128 \times 12)\), where $128$ represents the embedding size per each of the $12$ body parts, and a feedforward dimension of $3072$. Training is performed on a single H800 GPU through progressive conditioning: starting with text-only training for $460K$ steps, then adding reference motion for another $460K$ steps, followed by global spatiotemporal control for $230K$ steps, and finally incorporating full audio conditions for $920K$ steps. The corresponding batch sizes are $48$, $48$, $48$, and $16$, respectively. We optimize with AdamW using an initial learning rate of \(1 \times 10^{-4}\) (reset for new conditions) and a cosine schedule decaying to \(1 \times 10^{-5}\) within the first $460K$ steps. The default length of motion reference and prediction is $150$. 
During training and testing, the reference motion \(\mathbf{c}_{r}\) is provided by the preceding segment of the ground-truth sequence, which shares the same temporal length as the motion to be generated, or is set to null. At inference, this capability is extended to accommodate user-defined motions, enabling customized and interactive generation.
More implementation details are provided in supplementary material.

\subsection{Quantitative Results}
We evaluate \OurMethod~on four tasks: Text-to-Motion (T2M), Global Spatiotemporal Controllable Generation (GSTC), Music-to-Dance (M2D), and Speech-to-Gesture (S2G). 
%
For T2M and GSTC evaluations, test sets are uniformly sampled across all datasets~($10$ samples from each). 
The M2D benchmark integrates test sequences from AIST++~\cite{aistpp}, FineDance~\cite{finedance}, and PhantomDance~\cite{li2022danceformer}, with S2G evaluation conducted on BEAT2~\cite{liu2023emage}.

\textbf{T2M.}
Tab.~\ref{tab:T2M} shows that previous methods, constrained by small-scale datasets, struggle to generalize in generating diverse and complex motions. 
In contrast, \OurMethod~demonstrates a significant advantage, outperforming not only baseline methods trained on small datasets (MDM~\cite{mdm}, MLD~\cite{mld}, MoMask~\cite{momask}, MotionCraft~\cite{bian2024motioncraft}) but also MoMask* and MotionCraft*, which are trained on our \TrainingData. 
Compared to MoMask* and MotionCraft*, \OurMethod~uses stronger text encoder (T5-XXL~\cite{t5}), which models complex texts more effectively than CLIP~\cite{clip}.
In addition, it adopts a unified multimodal framework with reference motion during training, enabling the model to learn more detailed whole-body motion representations, leading to better generation quality and generalization.
We observed a significant performance enhancement when conditioning on the preceding ground-truth motion at test time, which validates the reference motion's role as a novel signal for preserving consistency in content, style, and temporal dynamics.

\textbf{GSTC.}
We adopt the cross-joint setup from OmniControl~\cite{omnicontrol}, simulating spatially dense control by controlling all joints. 
As shown in Tab.~\ref{tab:gstc}, due to the small dataset size, OmniControl~\cite{omnicontrol} struggles to generalize in GSTC task. 
In contrast, our method demonstrates clear advantages, effectively following spatially dense control signals.

\textbf{M2D and S2G.}
We directly compare our method with the MotionCraft~\cite{bian2024motioncraft} on the S2G and M2D tasks.
As shown in Tab.~\ref{tab:s2g_m2d}, \OurMethod~achieves competitive performance. The slightly higher FID is primarily because the test sets for the S2G and M2D tasks are relatively small and have a limited distribution. In contrast, \OurMethod~ is trained on \TrainingData, which comprises data from diverse tasks including T2M, M2D, S2G, HHI, HOI, and HSI. This diverse training endows our method with enhanced generative diversity, causing a distributional gap with the test sets of the S2G and M2D tasks.

\subsection{Ablation Study}

\begin{table}
  \centering
    \setlength{\tabcolsep}{1pt} 
    {\small 
    \begin{tabular}{lccccc}
      \toprule
        \multicolumn{1}{c}{\multirow{2}{*}{Task}} & \multicolumn{1}{c}{\multirow{2}{*}{Method}} &  
         \multicolumn{1}{c}{\multirow{2}{*}{FID$\downarrow$}} & \multicolumn{1}{c}{R Precision$\uparrow$} & \multirow{1}{*}{Multimodal } &  \multicolumn{1}{c}{\multirow{1}{*}{Diversity}}     \\ 
        &  &  & Top $3$ & Dist$\downarrow$ & $\rightarrow$ \\
      \midrule
      \multirow{2}{*}{\makecell{T2M}} & w/o TrSt &  \underline{9.574} & \underline{0.232} & \underline{6.853} & \underline{3.118}  \\ 
      & Ours & \textbf{5.040} & \textbf{0.571} & \textbf{4.678}  & \textbf{8.650}  \\ 
      \midrule
      \multirow{3}{*}{\makecell{GSTC}} &  w/o TrSt &   \underline{10.247} & \underline{0.491} & \underline{6.130} & \underline{2.438}  \\ 
      & Ours &   \textbf{4.224} & \textbf{0.686} & \textbf{4.377} & \textbf{6.292}  \\
      \bottomrule
    \end{tabular}
    }
    \caption{Ablation study on the weak-to-strong training strategy. \textbf{Bold} entries indicate the best results, and underlined entries indicate the second-best results.}
    \vspace{-5pt}
    \label{tab:ablation}
\end{table}
The ablation study in Tab.~\ref{tab:ablation} reveals two key findings: (1) Removing the progressive weak-to-strong conditioning training strategy compromises text-conditioned alignment, indicating that finer-grained controls can override coarse semantic constraints; (2) Mixed-condition training strategy impairs spatiotemporal control by introducing optimization conflicts due to physical constraints, which underscores the necessity of adopting a weak-to-strong conditioning strategy. Furthermore, our progressive training strategy not only resolves these conflicts but also significantly reduces computational overhead in the initial stages. This is attributed to its design, where multimodal conditions are introduced incrementally as training progresses, resulting in a lower computational load in the early phases. In contrast, a conventional end-to-end approach that jointly trains on all conditions from the outset converges more slowly and faces greater challenges in achieving convergence.

\subsection{Qualitative Results}

As shown in Fig.~\ref{fig:Diverse motion synthesis capabilities}, \OurMethod~handles various multimodal generation, including T2M, M2D, S2G, and GSTC (covering motion prediction, in-betweening, and joint guidance). When combined with reference motion, the model generates conditioned motions that align with the reference. More visualizations are in the supplementary video.

\section{Conclusion and Discussion}

This work introduces the \OurMethod~and the \TrainingData.
\OurMethod~is an autoregressive diffusion model that integrates reference motion and multimodal conditions for precise whole-body motion control. It employs a progressive weak-to-strong mixed conditions strategy to effectively handle multi-granular constraints.
\TrainingData~is the largest multimodal MoCap dataset, comprising $286.2$ hours from $28$ motion capture datasets, unified in SMPL-X with structured and consistent captions across 10 tasks. 
Experiments demonstrate that \OurMethod~outperforms baselines, laying a strong foundation for large-scale multimodal motion generation.

\textbf{Limitation and Future Work.} While our method demonstrates versatility across multimodal conditions, it has key limitations. A primary limitation is the absence of explicit conditioning for physical interactions with scenes, objects, or other persons. Consequently, the model struggles to generate plausible interactive motions in complex environments. Future work should thus focus on incorporating interaction-aware conditions, such as conditioning on scene geometry and enforcing explicit physical constraints, to enhance realism and versatility. Furthermore, inference speed remains a significant concern. Our model (27.22M parameters) requires 2.14 seconds per sample, which is slower than text-conditioned diffusion models of a comparable scale. Notably, this inference time only increases to 3.00s when scaling the model tenfold to 355.13M parameters. This indicates the primary bottleneck is not the model's parameters but the extensive context length arising from aggregating diverse multimodal conditions. Future research should therefore explore more efficient architectures for handling these long-sequence multimodal interactions to accelerate inference.

\section*{Acknowledgement}
This work was supported by the Science, Technology and Innovation Project of Shenzhen Longhua District (No. 20260309G23410662).

{
    \small
    \bibliographystyle{ieeenat_fullname}
    \bibliography{main}
}

\end{document}